\documentclass{article}

\usepackage{cite}
\usepackage{aaai18}
\usepackage{amsmath}
\usepackage{hyperref}
\usepackage{enumitem}
\title{Machine Learning in Cyber-Security - Problems, Challenges and Data Sets}
 \author{
    Idan Amit\textsuperscript{1}      , 
    John Matherly\textsuperscript{2}  ,
    William Hewlett\textsuperscript{1}   ,
    Zhi Xu\textsuperscript{1}         ,
    Yinnon Meshi\textsuperscript{1}   ,
    Yigal Weinberger\textsuperscript{1} \\
    \textsuperscript{1}{Palo Alto Networks} \\
    \textsuperscript{2}{Shodan}\\
    iamit@paloaltonetworks.com,
    jmath@shodan.io,
    whewlett@paloaltonetworks.com, \\
    zxu@paloaltonetworks.com, 
    ymeshi@paloaltonetworks.com, 
    yweinberge@paloaltonetworks.com}

\begin{document} 
\maketitle

\begin{abstract}
We present cyber-security problems of high importance. We show that in order to solve these cyber-security problems, one must cope with certain machine learning challenges. We provide novel data sets representing the problems in order to enable the academic community to investigate the problems and suggest methods to cope with the challenges.
We also present a method to generate labels via pivoting, providing a solution to common problems of lack of labels in cyber-security.
\end{abstract}
\section{Introduction}

Cyber-security is an important area in which machine learning is becoming increasingly significant. 
Many machine learning algorithms such as convolutional neural networks \cite{David2015}, LSTM \cite{Woodbridge2016} and others \cite{Khorshidpour2017,McLaughlin2017,Hu2017,Villegas2017,Patri2017} were applied to cyber-security problems.
It is important to note that machine learning in cyber-security is far more than merely applying established machine learning methods to data sets of cyber entities. 
 
Cyber-security involves machine learning challenges that require elegant methodological and theoretical handling. We believe that such challenges are of interest to people with passion for machine learning, without necessarily requiring cyber domain expertise or prior knowledge. The machine learning community is usually unaware of these challenges.
  
We are not the first to discuss security and AI challenges \cite{DBLP:journals/corr/abs-1712-05855} or alert on the lack of data sets  \cite{DBLP:journals/corr/abs-1709-07095}. The novelty of our work is the presentation of new cyber-security problems, the machine learning challenges involved in them and the publication of data sets that enable investigating them. We hope it will lead to new methods in both machine learning and cyber-security.

\section{Cyber-Security Problems}

\subsection{Malware Classification And Detection : Identifying Malicious Programs}
Malware is a program or a file that is harmful to a computer system.
As part of the arms race, the attacker tries to avoid detection.
Some evasion techniques are polymorphism, impersonation, compression and obfuscation \cite{5633410}. For example, in malware coloring the attacker slightly change the malware, leading to polymorphism (many variants).  
Since many threat intelligence repositories are based on signatures of the malware (e.g., SHA1, MD5), a slight modification enables to a signature-based detection.

Current detection systems use various algorithms: Naive Bayes Classifier\cite{Luo2016}, SVM \cite{Kuriakose2015}, Random Forest \cite{Hu2017a}, DNN \cite{Xie2015}, CNN \cite{David2015} and LSTM \cite{Woodbridge2016,Saxe2017}.  

\subsubsection{Labeling Malware Via Opertor Domain Pivoting}
Historically, malware classification was based on signatures and domain experts.
However, domain experts are limited in the number of cases they analyze, and signatures lead to labeling errors.
The large number of malware and their population rapid growth make the problem even more severe.
For these reasons, we have constructed a malware data set and grouped malware contacting the same malicious site.

Let $m_{1},m_{2}$ be malware.

Let $OperatorDomains(m)$ be the unique domains with which the malware communicates. 

The specific definition of $OperatorDomains(m)$ is use case specific and requires domain knowledge.
For example, many malware communicate with benign domains (e.g., google.com) in order to check Internet connectivity. So, communication with Google is not an indication of the operator or maliciousness.
A good definition of $OperatorDomains(m)$ should focus in malicious (or at least not known benign domains) that are not communicating with too many files.

If $OperatorDomains(m_{1}) \cap OperatorDomains(m_{2}) \neq \emptyset$, then $m_{1},m_{2}$ belong to the same operator.

Hence, given a data set $D$ of malware and a $OperatorDomains(m)$ function we label $	\forall m_{1},m_{2} \in D, (m_{1},m_{2}, OperatorDomains(m_{1}) \cap OperatorDomains(m_{2}) \neq \emptyset)$.

In some uses it is more appropriate to work with a multi-class data set, assigning to each pair an indication of the class the pair belongs to (e.g., the mutual domain).

Thanks to that, we have a large number of labels, agnostic to the content, that can be used to build and evaluate content-based malware classification models. To our knowledge, this is the first time that a domain-based grouping data set is suggested and provided.

\subsection{Host Similarity : Identifying Malicious Sources}
The same malicious operator will tend to use hosts with the same services on them, since they are used for the same need. Having a similarity function among hosts in the Internet, it will enable us to deduce on one host using similar hosts. Assume that we see two hosts that use the same exact version of MySQL, PHP and many other products. It is unlikely to be coincidence and if there is threat intelligence on malicious activity of one host, then it is likely the other host belongs to the same operator and is used for malicious activity too. Like in the malware analysis, we can use malware in order to label related hosts. If a malware communicates with two malicious domains, the malware are likely to belong to the same operator and so are the domains. Given the services profile of the hosts, we can learn the similarity function and use it to identify new domains for which we didn't see malware communication. 
Attributing different domains to the same operator was done before \cite{Starov:2018:BYD:3178876.3186089} but as far as we know we are the first to present host similarity problem based on agnostic label, and provide a data set for it.

\subsubsection{Labeling Hosts Via Operator Domain Pivoting}

The labeling method for hosts is based on the malware domain labeling.
For each domain $d$, one can resolve the $ip$ address on which it is being hosted (e.g., via nslookup). We will mark the resolving function as $resolve(ip)$.
Shodan provide scans of hosts in the Internet, giving the $signature(ip)$ of services available on the host.
Hence, given a data set $D$ of malware , and the functions $OperatorDomains(m),resolve(ip), signature(ip)$ function we label 

$	\forall m_{1},m_{2} \in D, \forall d \in OperatorDomains(m_{1}) \cap OperatorDomains(m_{2}),$

$\forall ip_{1}, ip_{2} \in resolve(d)$
$(signature(ip_{1}), signature(ip_{2}), True)$.

The negative samples are due to a Cartesian product of signatures of ip address that don't obey the positive samples rule.

\subsection{Lateral Movement : The Attacker Moves to its Target}
An attacker usually starts an attack in a point in the network that is far from the final destination. Therefore, the attacker should learn the network and travel to the destination, a phase called lateral movement \cite{6578801}. This path might be long and use edges that on their own might be reasonable but as a part of a path they might be very suspicious. The goal of the defender is to learn the access pattern graphs (which might differ by usage) and use it in order to assign probabilities to paths.

\subsection{Stealth Port Scan : The Attacker Explores its Target}

An attacker that has a hold on one host (computer) in a network is likely to want to gain a hold on other hosts too. Knowing which services are deployed on a host might help the attacker to use an exploit for one of them. Accessing common ports related to these services can help with mapping the available services. A port scan \cite{4772622} is access to a sequence of ports in order to map them. This technique is problematic from the point of view of the attacker since a scan of many ports is very noisy and the attacker might be detected. More sophisticated attackers, called Advanced Persistent Threat (APT) \cite{6657248}, use stealth port scans \cite{Staniford:2002:PAD:597917.597922, Aniello:2011:ISP:1978582.1978597, Streilein2001ImprovedDO}. They select very few informative ports in order to profile the target host. A possible method to identify a stealth port based on the low probability of access to a given combination of ports. 

\subsection{Bind and Reverse Shell : The Attacker Gets a Hold on its Target}
As part of a manually operated targeted attack process, the attacker may be advancing from host A to host B, ultimately gaining a remote interactive command shell on host B.
Two very common approaches to achieve a remote shell are reverse shell and bind shell, and they can be used interchangeably, commonly depending on the attacked network structure and firewall configurations. Both methods require two consecutive connections between the source and the destination and are available in common penetration tools such as Metasploit (\url {https://www.metasploit.com/}).
In the bind shell scenario, an initial connection from A to B is used to exploit the vulnerability on a specific port (compromised service) and a follow-up connection from A to B is used as an interactive shell on a different (and commonly unused by other services) port. In the reverse shell scenario, the follow-up connection is from B to A and not from A to B. To our knowledge, this is the first time that a labelled data set is provided for this problem.

\section{Machine Learning Challenges in Cyber-Security}

We detail the below machine learning challenges in cyber-security. These challenges were chosen since they are common in cyber-security problems and that coping with them is essential for solution. For example, imbalanced data sets appear whether one tries to do malware detection, domain reputation or identify network intrusion. If one fails to identify an imbalanced data set, one might end up with a model claiming: "Everything is benign". While the model will have astonishing accuracy, it will provide no value. Off course, the imbalanced challenge is not unique to the cyber security domain, yet it is an essential challenge in many cyber security problems. 

\subsection{Attacker-Defender Game}  

Cyber-security contexts may have different, purposeful adversaries with different aims, techniques of attack, and capabilities, the latter including the knowledge of their adversary (the defender). Advancement of one of the sides will not end the game but will lead to a new round with different settings. 
Citing Slick Willie Sutton: "I rob banks because that’s where the money is"; We know that this game will be played for long time.

Game theory has been applied to cyber-security \cite{Manshaei:2013:GTM:2480741.2480742}, \cite{Roy2010ASO}.
What makes the game more interesting is the asymmetries in it. The defense systems are usually deployed before the attack and the attacker can treat them as given. We can assume that the attacker knows which defense system they should cope with. The attacker might even have access to the defense system and know its logic.

There are also asymmetries in the knowledge. The attacker knows the goals of the attack, the timing and the attack steps. The defender is familiar with the protected assets and the attacked network. A typical example of the use of such asymmetry in the knowledge is the use of honeypots \cite{1254322}.
Honeypots are traps, hosts that no benign user should access. The attacker, not knowing that, might access the honeypot and found by the defender. 

The defender can win not only by preventing an attack, but also by making it infeasible. Cyber-espionage is less attractive if the secrets will be found out years after they were used. Cyber-crime economics will make long attack not profitable in most cases.

The last interesting point is considering not a single attack as a game but the entire cyber-security arena as a game. A malware caught in one network might be uploaded to common threat intelligence repositories (e.g. \url{https://www.virustotal.com/}) and be identified in other locations. The defending side can proactively hunt resources and tools used by attackers (e.g., malicious domains, files signatures), making the compromising of future attacks easier. Coping with the host similarity problem is an important step in this direction.

\subsection{Lack of Labeled Samples and Certainty in Ground Truth}   

Most cyber-security tasks are essentially supervised learning tasks. Given an entity or activity, we should decide whether it is malicious or benign. Unfortunately, many times we lack the labels which are required for supervised learning.

Manual labeling is limited in scope and subjective. Therefore, most labels are coming from heuristics. Other than having errors, one might end up trying to model the heuristics he started with.

Using the community knowledge is a common approach. When the entity to classify is universal (appearing also outside the examined case, like a malware or a malicious domain but not network activity), one can use external black lists and white lists for labels. This method may introduce a domain adaption problem , since the lists are not coming from the same source as the entities to classify. 
Off course, such lists might be of low quality.

Consensus agreement, using the verdict of many vendors as the concept, is another common solution. Other than the potential intellectual property problems, there are algorithmic problems with this approach. The consensus doesn't necessarily agree with the ground truth. Important demonstration of this problem is with Advanced Persistent Threats (APT) \cite{6657248}. Attacks of APTs are usually sophisticated and targeted. An APT will tend to use its malware on a single target and therefore its malware won't have threat intelligence recognizing it and so the malware will evade detection. Using consensus agreement as the concept is like asking your model not to detect APTs.

Algorithmic solutions might involve the use of unsupervised learning \cite{Hastie2009} methods and specifically anomaly detection \cite{Chandola:2009:ADS:1541880.1541882}. 
While anomaly detection is popular in many domains, it is usually not used in cyber-security due to inherent difficulties \cite{5504793}, that we see as machine learning challenges.
These solutions may be problematic since some domains, like networks, have a plethora of benign anomalies. A sophisticated attacker is likely to be aware of the defender and try to make sure that its actions do not look anomalous. In this scenario we might end up with plenty of anomalies, none of which is related to the attacker.

Some promising research directions are active learning \cite{Settles10activelearning} and semi-supervised learning \cite{Zhu06semi-supervisedlearning}. Active learning can help in identifying the most informative entities to label, and help in building a small labeled data set. One should note that some of the labeling is subjective and in many cases it is not possible to label with high certainty. This leads us to learning in the presence of labeling errors \cite{Kearns93learningin}, \cite{Kearns:1998:ENL:293347.293351}.
Given a small labeled data set, one can apply semi supervised learning and by doing so be able to predict as in supervised learning.

\subsection{Imbalanced Data Sets}     

We consider a data set to be imbalanced when the ratios between the majority set and the minority sets are large \cite{Chawla04editorial:special}. While that in the machine learning community a ratio of 1 to 10 is considered imbalanced, malicious training examples in cyber-security data sets may be extremely rare, and imbalance ratios of 1 to 10,000 are common.      

When the ratio between the sets is large but there are enough samples in each set, we use the term relative imbalance. In the more severe case in which there are not enough samples in the minority set, this is absolute imbalance. This is a common situation in cyber-security.

\subsection{The Tragedy of Metrics}   

As we said, in the typical cyber-security use case we have only a handful positive labeled samples and we usually don't know how many other samples in our data set are actually positive.

This leads to a tragedy that goes much further than algorithmic considerations.
Since we don't know what the actual positives are, we usually cannot know the positive rates. This in turn means that we cannot estimate our recall.
Note that the cost of a false positive is a waste of some hours, loss of confidence in the system and in the long term the loss of a customer by the vendor. On the other hand, the cost of false negative might be millions of times more expensive.
The inability of estimating recall leads customers and vendors into improving the precision. Since it is easy to trade off recall for precision by requiring higher confidence, we might expose the protected assets to more risk by racing to reduce false alarms.

A common method to handle this scenario is to use a cost matrix and assign a higher cost to false negatives. This method is not directly applicable here since we are not aware of our false negatives.

Achieving high lift is usually passed unnoticed.  If the positive ratio is 1 to 10,000 and our classifier’s precision is 10\% than our precision lift is 1,000. However, the user of the system observes a single success in 10 alerts, which is not observed as good performance.

The common assumption of Independent Identical Distribution doesn't hold in cyber-security. For example, malware coloring leads to many polymorphic variations of the same malware. These instances are very dependent, making the usual statistical guarantees misleading. Even without deliberate attempts the assumption might be violated. Hosts in the same network segment (e.g., DMZ) are related. Being able to identify hardware of the same vendor will probably won't generalize well to other vendors. There is a need for mathematical methods to cope with these scenarios.

\subsection{Domain Adaptation}  

Domain adaptation is a scenario in which the test distribution on which a model is evaluated is different from the train distribution that was used to build the model \cite{DBLP:journals/corr/abs-0907-1815}. The cyber world looks different in each of its areas. Threats, entities and networking behavior are different among parties acting in various scales, domains, cultures and geo-locations. A basic challenge therefore is adapting an effective defense from one domain to another.

\subsection{Concept Drift}   

A security expert copes successfully with the challenges listed above and deploys high performing system to production environment.  
In a short time, the system suffers a severe degradation in performance. The model did not change. It is the world that changed.
Cyber-security is a rapidly evolving field. Models and insights are likely to become obsolete quickly.  Volumes change due to technological advancement, new protocols and domains appear and malicious activity also has trends and fashions. The change in the source of analyzed entity due to a change in time is called concept drift \cite{GU201329}. To avoid continually starting from scratch, ways to track concept drift and extreme changes in the domain need to be developed.    

\section{Data Sets}
We think that the use of machine learning in cyber-security should change. We also think that the cyber community should help the machine learning community to become more involved in this field.   

One of the key obstacles to investigating cyber-security problems is the lack of appropriate data sets. There are many important cyber-security data sets like Microsoft's malware data set \cite{DBLP:journals/corr/abs-1802-10135}, Los Alamos's traffic data set \cite{turcotte} and EndGame's Ember malware properties data set \cite{DBLP:journals/corr/abs-1804-04637}. However, we feel that there are no suitable data sets that will enable academic researchers to cope with the problems and challenges we listed.

In general, companies do not tend to share their data, and cyber-security data is even more sensitive than usual. However, by using anonymization, a bit older data, and removing sensitive information, one can address most concerns. Once this is done, the data contributing company will be able to enjoy from the combined work of a world of experts working on cyber problems.   

The following data sets were contributed by Palo Alto Networks and Shodan (for academic use only). We do hope that these data sets will boost the research of machine learning in cyber-security. We also hope that other organizations will follow and publish more data sets and the research will be further boosted.

\subsection*{Access to Data Sets}

We would like to provide researchers access to the data sets for academic use. In order to gain access to the data sets please contact data-sets@paloaltonetworks.com with "Access to data request" in the title.

\subsection{Malware Polymorphism}

We relate a malware to the domains it contacted.
The data set is made of a sample of malware identified by Palo Alto Networks in a given period.
For each malware we provide identifiers and the domains accessed by the file.
One can find malware variants by creating pairs of files communicating with the same domains.
Examples of pairs that are not variants can be generated by matching files communicating with no common domain.
Note that one can generate $O(n^2)$ negative pairs and it is up to the researcher to choose the threshold that fits best its needs.

The malware data set construction demonstrates away to cope with the lack of labeled samples. The labels are imbalanced.
The variety of malware lead to domain adaptation.
We would like to provide in the future data collected on later periods, and present concept drift.

Malware data set
\begin{itemize}
  \item $SHA\-256$ The signature of the file given the SHA-256 hash function.
  \item $md5$ The signature of the file given the MD5 hash function.
  \item $ssdeep$ The signature of the file given the SSDEEP fuzzy hash function
  \item $size$ File size
\end{itemize}

Communication data set
\begin{itemize}
  \item $SHA\-256$ The signature of the file given the SHA-256 hash function.
  \item $domain$ The name of the domain with which the malware communicated. A malware that communicated with few domains will have a record per domain.
  \item $IP$ The IP address to which the domain was resolved.
\end{itemize}

\subsection{Host Similarity}
The host similarity data set is based on the malware polymorphism data set.
We consider two hosts (represented as IP addresses) as similar if the same malware communicated with both. 

The host similarity data set construction demonstrates a way to cope with the lack of labeled samples. The labels are imbalanced.
The variety of host lead to domain adaptation.
We would like to provide in the future data collected on later periods, and present concept drift.

Shodan (\url{https://www.shodan.io/}) contributed the host services signature:
The list of services opened on the host and the profile of each service. For example, see \url{https://www.shodan.io/host/180.183.160.140}.
Given the pairs of related and unrelated host, generated by malware polymorphism, one can learn a similarity function in the services space. 
This similarity function enables identifying hosts similar to a host in question. Given that the similar hosts are malicious, it is likely that the considered host belongs to the same operator and is also malicious.

The communication data set from the malware polymorphism section is applicable here too.

For each IP address, we provide the the scan result in a JSON format. For details about the format, see Shodan \url{https://www.shodan.io/}).

\subsection{Ngrams Data Set}

The data set is a derived from a series of benign and malicious files, with identifying names (SHA256s) removed, gathered over a period of about a month in 2017.  For each file, we capture a histogram of 4-grams of the byte code. For example, the string 01234 would produce the 4-grams 0123 and 1234.  Note that you can derive approximate (exact except for the end of file) 1, 2, and 3 grams from this data.  We also have labels and family tags. The labels are benign, malicious, questionable. The tags describe families of malware, not every sample has a family tag, some have more than one. The median number of 4grams per file is 105,888. 

The variety of malware lead to domain adaptation.
We would like to provide in the future data collected on later periods, and present concept drift.

There are two pieces of data:
\begin{enumerate}
\item A mapping of files indices to ngrams where each line is a tab separated list:
\begin{itemize}
\item File Index
\item Ngram 1 : Ngram 1 Count
\item Ngram 2 : Ngram 2 Count
\item etc.
\end{itemize}
\item A mapping of files to labels where each line is a tab separated list:
\begin{itemize}
\item File Index
\item Verdict (0=benign, 1=malware, 2=greyware)
\item Family Label 1
\item Family Label 2
\item etc.
\end{itemize}
\end{enumerate}

\subsection{Android Malware Family Data Set}

The data set is a derived from a series of mixed Android APK files, with identifying names (SHA256s) removed, gathered over a period of about a month in 2017. For each file, we provide an XML report of metadata generated during the static and dynamic analysis (generated by $WildFire A^{TM}$. For example, the certificate that used to sign the Android APK file; the domains that being contacted after the APK file got installed; the API call sequences after installation.  We also have labels and family tags. The labels are benign, malicious, greyware. The family tags describe family names of malware. One malware may have multiple family tags.  

The labels in this data set are imbalanced.
The variety of malware lead to domain adaptation.
We would like to provide in the future data collected on later periods, and present concept drift.

There are two pieces of data:
\begin{enumerate}
\item A mapping of files to labels and tags:
\begin{itemize}
\item File Index
\item static\_report : a JSON object
\item dynamic\_report : a JSON object
\item file\_info : a JSON object
\end{itemize}
\item A mapping of files to labels where each line is a tab separated list:
\begin{itemize}
\item File Index
\item Verdict (0=benign, 1=malware, 2=greyware)
\item Family Label 1
\item Family Label 2
\item etc.
\end{itemize}
\end{enumerate}

\subsection{Bind Shell Data Set}

The bind shell data set was collected from some networks over a period of 4 months. It consists of a set of consecutive connection pairs, that were created by pairing connections from source to destination that follow the following rules:

The data set presents the challenge of lack of labels and is also highly imbalanced. 

\begin{enumerate}
\item Both connections are observed within a specific short time frame.
\item The destination port of the first connection is different from the second connection.
\item The connection pairs make good candidates for actual bind shell attack, based on a noise filtering algorithm applied at collection time. \end{enumerate}

The aggregative features mentioned below that relate to counts of hosts and ports are calculated on the connection pairs from the entire network without noise filtering over a given time period.

Each connection pair has the following features:
\begin{itemize}
\item index ,  Index of the sample
\item label ,  Is forward shell, 1 for true, 0 for false, -1 for missing
\item source\_host\_id ,  Index of the source
\item is\_new ,  True if the source and destination hosts were not seen on the network for a period of time prior to the creation of the connection pair
\item s\_phase1\_initiators\_hosts ,  Amount of Destinations accessed from this source using the same phase\_1 port
\item s\_phase2\_initiators\_hosts ,  Amount of destinations accessed from this source using the same phase\_2 port
\item s\_phase1\_initiators\_ports ,  Amount of distinct phase\_2 ports accessed from this source to all destinations using the same phase\_1 ports
\item s\_phase2\_initiators\_ports ,  Amount of distinct phase\_1 ports accessed from this source to all destinations using the same phase\_2 port
\item s\_port\_count ,  Popularity of this phase\_1 and phase\_2 ports couple across the network
\item s\_src\_port\_phase1 ,  Port used in the first session
\item s\_src\_port\_phase2 ,  Port used in the second session
\item s\_pair\_phase1\_cnt ,  Amount of distinct phase\_1 ports between this source and destination
\item s\_pair\_phase2\_cnt ,  Amount of distinct phase\_2 ports between this source and destination
\item s\_start\_time\_phase1 ,  Start time of the first session
\item s\_start\_time\_phase2 ,  Start time of the second  session
\item s\_duration\_phase1 ,  Duration of the first session
\item s\_duration\_phase2 ,  Duration of the second  session
\item s\_dst\_port\_phase1 ,  Destination port of the first session 
\item s\_dst\_port\_phase2 ,  Destination port of the second  session 
\item s\_volume\_phase1 ,  Volume (source to destination) of the first session
\item s\_volume\_phase2 ,  Volume of the second session
\item s\_rvolume\_phase1 ,  Returning volume (destination to source) of the first session
\item s\_rvolume\_phase2 ,  Returning volume (destination to source) of the second session
\item s\_path\_phase1 ,  Protocol used in the first session
\item s\_path\_phase2 ,  Protocol used in the second session
\item s\_spfss\_unique\_srcs ,  Amount of distinct sources creating connections to the candidate's destination with the same phase\_1 and phase\_2 ports
\item s\_arb\_host\_count ,  Amount of sources that accessed this destination with the same phase\_1 port and followed up with another connection (i.e. a phase\_2 session was detected)
\item s\_arb\_port\_count ,  Amount of distinct phase\_2 ports accessed after accessing this destination and phase\_1 port
\end{itemize}

\subsection{Network Traffic Data Set}

The network traffic data set is a collection of network sessions.
The sessions are aggregated in 10 minutes buckets. Hence, if A communicated with B twice in that bucket there will be a single traffic record with cnt=2.

The data set lacks labels and even the concepts of stealth port scan and lateral movement are a bit vague (e.g., How long should be a path of hosts in order to be considered as lateral movement).
The data sets were collected from different sites, presenting domain adaptation challenge.
Data set from different periods present concept drift.

The columns are the following:
\begin{itemize}
  \item $min\_start\_time$ Time of the session, data set beginning is set to Alan Turing’s birthday, June 23th, 1912 , and relative time is given.
  \item $src\_index$ Source IP representation. IP addresses are collected in this in the index in the data set (in order to keep privacy).
  \item $dst\_index$ Destination IP representation. IP addresses are collected in this in the index in the data set (in order to keep privacy).
  \item $src\_port$ Source port
  \item $dst\_port$ Destination port
  \item $tvolume$ Volume of traffic from source to destination
  \item $rtvolume$ Volume of traffic from destination to source
  \item $pkt$ Number of packets from source to destination
  \item $rpkt$ Number of packets destination to source
  \item $cnt$ Number of sessions
  \item $failed\_num$ number of failed sessions
  \item $path$ Protocol used for communication
\end{itemize}

This data set can be used for the cyber problems of stealth port scan and lateral movement.

\bibliographystyle{abbrv}
\bibliography{bibtex.bib}

\end{document}